\documentclass{article}

 \PassOptionsToPackage{square,numbers,sort&compress}{natbib}
\usepackage[preprint]{neurips_2019_arxiv}

\usepackage[utf8]{inputenc} % allow utf-8 input
\usepackage[T1]{fontenc}    % use 8-bit T1 fonts
\usepackage{hyperref}       % hyperlinks
\usepackage{url}            % simple URL typesetting
\usepackage{booktabs}       % professional-quality tables
\usepackage{amsfonts}       % blackboard math symbols
\usepackage{nicefrac}       % compact symbols for 1/2, etc.
\usepackage{microtype}      % microtypography
\usepackage{amsmath}  % Custom
\usepackage{enumitem}  % Custom
\usepackage{pdfpages} % Custom - for appendix
\usepackage{subcaption} 
\usepackage{pifont}
\usepackage{wrapfig}

\newtheorem{theorem}{Theorem}
\DeclareMathOperator{\E}{\mathbb{E}}

\usepackage{algorithm}
\usepackage{algorithmicx}
\usepackage[noend]{algpseudocode}

\newcommand{\isep}{\mathrel{{.}\,{.}}\nobreak}

\title{Perfect Match:\\ A Simple Method for Learning Representations For Counterfactual Inference With Neural Networks}

\author{
  Patrick Schwab$^{1}$, Lorenz Linhardt$^{2}$, Walter Karlen$^{1}$ \\
  $^{1}$Institute of Robotics and Intelligent Systems, ETH Zurich, Switzerland \\
  $^2$ Department of Computer Science, ETH Zurich, Switzerland\\
  \texttt{patrick.schwab@hest.ethz.ch} \\
}

\begin{document}

\maketitle
\vspace{-3ex}

\begin{abstract}
Learning representations for counterfactual inference from observational data is of high practical relevance for many domains, such as healthcare, public policy and economics. Counterfactual inference enables one to answer "What if...?" questions, such as "What would be the outcome if we gave this patient treatment $t_1$?". However, current methods for training neural networks for counterfactual inference on observational data are either overly complex, limited to settings with only two available treatments, or both. Here, we present Perfect Match (PM), a method for training neural networks for counterfactual inference that is easy to implement, compatible with any architecture, does not add computational complexity or hyperparameters, and extends to any number of treatments. PM is based on the idea of augmenting samples within a minibatch with their propensity-matched nearest neighbours. Our experiments demonstrate that PM outperforms a number of more complex state-of-the-art methods in inferring counterfactual outcomes across several benchmarks, particularly in settings with many treatments.
\end{abstract}
\vspace{-3ex}

\section{Introduction}
Estimating individual treatment effects\footnote{The ITE is sometimes also referred to as the conditional average treatment effect (CATE).} (ITE) from observational data is an important problem in many domains. In medicine, for example, we would be interested in using data of people that have been treated in the past to predict what medications would lead to better outcomes for new patients \cite{shalit2016estimating}. Similarly, in economics, a potential application would, for example, be to determine how effective certain job programs would be based on results of past job training programs \cite{lalonde1986evaluating}. 

ITE estimation from observational data is difficult for two reasons: Firstly, we never observe all potential outcomes. If a patient is given a treatment to treat her symptoms, we never observe what \textit{would have happened if} the patient was prescribed a potential alternative treatment in the same situation. Secondly, the assignment of cases to treatments is typically biased such that cases for which a given treatment is more effective are more likely to have received that treatment. The distribution of samples may therefore differ significantly between the treated group and the overall population. A supervised model na\"ively trained to minimise the factual error would overfit to the properties of the treated group, and thus not generalise well to the entire population. 

To address these problems, we introduce Perfect Match (PM), a simple method for training neural networks for counterfactual inference that extends to any number of treatments. PM effectively controls for biased assignment of treatments in observational data by augmenting every sample within a minibatch with its closest matches by propensity score from the other treatments. PM is easy to use with existing neural network architectures, simple to implement, and does not add any hyperparameters or computational complexity. We perform experiments that demonstrate that PM is robust to a high level of treatment assignment bias and outperforms a number of more complex state-of-the-art methods in inferring counterfactual outcomes across several benchmark datasets. The source code for this work is available at \url{https://github.com/d909b/perfect\_match}. 

\noindent\textbf{Contributions.} This work contains the following contributions:
\begin{itemize}[noitemsep,leftmargin=2.2ex]
\item We introduce Perfect Match (PM), a simple methodology based on minibatch matching for learning neural representations for counterfactual inference in settings with any number of treatments. 
\item We develop performance metrics, model selection criteria, model architectures, and open benchmarks for estimating individual treatment effects in the setting with multiple available treatments.
\item We perform extensive experiments on semi-synthetic, real-world data in settings with two and more treatments. The experiments show that PM outperforms a number of more complex state-of-the-art methods in inferring counterfactual outcomes from observational data.
\end{itemize}

\section{Related Work}
\label{sec:related_work}
\paragraph{Background.} Inferring the causal effects of interventions is a central pursuit in many important domains, such as healthcare, economics, and public policy. In medicine, for example, treatment effects are typically estimated via rigorous prospective studies, such as randomised controlled trials (RCTs), and their results are used to regulate the approval of  treatments. However, in many settings of interest, randomised experiments are too expensive or time-consuming to execute, or not possible for ethical reasons \cite{carpenter2014reputation,bothwell2016rct}. Observational data, i.e. data that has not been collected in a randomised experiment, on the other hand, is often readily available in large quantities. In these situations, methods for estimating causal effects from observational data are of paramount importance.

\paragraph{Estimating Individual Treatment Effects.} Due to their practical importance, there exists a wide variety of methods for estimating individual treatment effects from observational data. However, they are predominantly focused on the most basic setting with exactly two available treatments. Matching methods are among the conceptually simplest approaches to estimating ITEs. Matching methods estimate the counterfactual outcome of a sample $X$ with respect to treatment $t$ using the factual outcomes of its nearest neighbours that received $t$, with respect to a metric space. These k-Nearest-Neighbour (kNN) methods \cite{ho2007matching} operate in the potentially high-dimensional covariate space, and therefore may suffer from the {curse of dimensionality} \cite{indyk1998approximate}. Propensity Score Matching (PSM) \cite{rosenbaum1983propensity} addresses this issue by matching on the scalar probability $p(t|X)$ of $t$ given the covariates $X$. Another category of methods for estimating individual treatment effects are adjusted regression models that apply regression models with both treatment and covariates as inputs. Linear regression models can either be used for building one model, with the treatment as an input feature, or multiple separate models, one for each treatment \cite{kallus2017recursive}. More complex regression models, such as Treatment-Agnostic Representation Networks (TARNET) \cite{shalit2016estimating} may be used to capture non-linear relationships. Methods that combine a model of the outcomes and a model of the treatment propensity in a manner that is robust to misspecification of either are referred to as {doubly robust} \cite{funk2011doubly}. Tree-based methods train many weak learners to build expressive ensemble models. Examples of tree-based methods are Bayesian Additive Regression Trees (BART) \cite{chipman2010bart,chipman2016bayestree} and Causal Forests (CF) \cite{wager2017estimation}. Representation-balancing methods seek to learn a high-level representation for which the covariate distributions are balanced across treatment groups. Examples of representation-balancing methods are Balancing Neural Networks \cite{johansson2016learning} that attempt to find such representations by minimising the {discrepancy distance} \cite{mansour2009domain} between treatment groups, and Counterfactual Regression Networks (CFRNET) \cite{shalit2016estimating} that use different metrics such as the Wasserstein distance. Propensity Dropout (PD) \cite{alaa2017deep} adjusts the regularisation for each sample during training depending on its treatment propensity. Generative Adversarial Nets for inference of Individualised Treatment Effects (GANITE) \cite{yoon2018ganite} address ITE estimation using counterfactual and ITE generators. GANITE uses a complex architecture with many hyperparameters and sub-models that may be difficult to implement and optimise. Causal Multi-task Gaussian Processes (CMGP) \cite{alaa2017bayesian} apply a multi-task Gaussian Process to ITE estimation. The optimisation of CMGPs involves a matrix inversion of $O(n^3)$ complexity that limits their scalability. 

In contrast to existing methods, PM is a simple method that can be used to train expressive non-linear neural network models for ITE estimation from observational data in settings with any number of treatments. PM is easy to implement, compatible with any architecture, does not add computational complexity or hyperparameters, and extends to any number of treatments. While the underlying idea behind PM is simple and effective, it has, to the best of our knowledge, not yet been explored. 

\section{Methodology}
\paragraph{Problem Setting.} We consider a setting in which we are given $N$ i.i.d. observed samples $X$, where each sample consists of $p$ covariates $x_i$ with $i\in[0 \isep p-1]$. For each sample, the potential outcomes are represented as a vector $Y$ with $k$ entries $y_j$ where each entry corresponds to the outcome when applying one treatment $t_j$ out of the set of $k$ available treatments $T = \{t_0,..., t_{k-1}\}$ with $j\in[0 \isep k-1]$. As training data, we receive samples $X$ and their observed factual outcomes $y_j$ when applying one treatment $t_j$, the other outcomes can not be observed. The set of available treatments can contain two or more treatments. We refer to the special case of two available treatments as the binary treatment setting. Given the training data with factual outcomes, we wish to train a predictive model $\hat{f}$ that is able to estimate the entire potential outcomes vector $\hat{Y}$ with $k$ entries $\hat{y}_j$. In literature, this setting is known as the Rubin-Neyman potential outcomes framework \cite{rubin2005causal}. 

\paragraph{Assumptions.} Counterfactual inference from observational data always requires further assumptions about the data-generating process \cite{pearl2009causality,peters2017elements}. Following \cite{imbens2000role,lechner2001identification}, we assume unconfoundedness, which consists of three key parts: (1) Conditional Independence Assumption: The assignment to treatment $t$ is independent of the outcome $y_t$ given the pre-treatment covariates $X$, (2) Common Support Assumption: For all values of $X$, it must be possible to observe all treatments with a probability greater than 0, and (3) Stable Unit Treatment Value Assumption: The observed outcome of any one unit must be unaffected by the assignments of treatments to other units. In addition, we assume smoothness, i.e. that units with similar covariates $x_i$ have similar potential outcomes $y$.

\paragraph{Precision in Estimation of Heterogenous Effect (PEHE).} The primary metric that we optimise for when training models to estimate ITE is the PEHE \cite{hill2011bayesian}. In the binary setting, the PEHE measures the ability of a predictive model to estimate the difference in effect between two treatments $t_0$ and $t_1$ for samples $X$. To compute the PEHE, we measure the mean squared error between the true difference in effect $y_1(n) - y_0(n)$, drawn from the noiseless underlying outcome distributions $\mu_1$ and $\mu_0$, and the predicted difference in effect $\hat{y}_1(n) - \hat{y}_0(n)$ indexed by $n$ over $N$ samples:
\begin{align}
\label{eq:PEHE} \epsilon_\text{PEHE} = \frac{1}{N}\sum_{n=0}^{N}&\Big( \mathbb{E}_{y_j(n)\sim\mu_j(n)}[y_1(n) - y_0(n)] - [\hat{y}_1(n) - \hat{y}_0(n)] \Big)^2
\end{align}
When the underlying noiseless distributions $\mu_j$ are not known, the true difference in effect $y_1(n) - y_0(n)$ can be estimated using the noisy ground truth outcomes $y_i$ (Appendix A). As a secondary metric, we consider the error $\epsilon_{\text{ATE}}$ in estimating the average treatment effect (ATE) \cite{hill2011bayesian}. The ATE measures the average difference in effect across the whole population (Appendix B). The ATE is not as important as  PEHE for models optimised for ITE estimation, but can be a useful indicator of how well an ITE estimator performs at comparing two treatments across the entire population. We can neither calculate $\epsilon_\text{PEHE}$ nor $\epsilon_\text{ATE}$ without knowing the outcome generating process.

\paragraph{Multiple Treatments.} Both $\epsilon_\text{PEHE}$ and $\epsilon_\text{ATE}$ can be trivially extended to multiple treatments by considering the average PEHE and ATE between every possible pair of treatments. Note that we lose the information about the precision in estimating ITE between specific pairs of treatments by averaging over all ${k \choose 2}$ pairs. However, one can inspect the pair-wise PEHE to obtain the whole picture.
\noindent\begin{minipage}{.5\linewidth}
\begin{small}
\begin{align}
\label{eq:multi_PEHE} \hat\epsilon_\text{mPEHE} &= \frac{1}{{k \choose 2}}\sum_{i=0}^{k-1}\sum_{j=0}^{i-1} \hat\epsilon_{\text{PEHE},i,j}
\end{align}
\end{small}
\end{minipage}%
\begin{minipage}{.5\linewidth}
\begin{small}
\begin{align}
\label{eq:multi_ATE} \hat\epsilon_\text{mATE} &= \frac{1}{{k \choose 2}}\sum_{i=0}^{k-1}\sum_{j=0}^{i-1} \hat\epsilon_{\text{ATE},i,j}t
\end{align}
\end{small}
\end{minipage}

\begin{wrapfigure}{r}{0.55\textwidth}
\vskip -1ex
\centering
\includegraphics[width=0.65\linewidth]{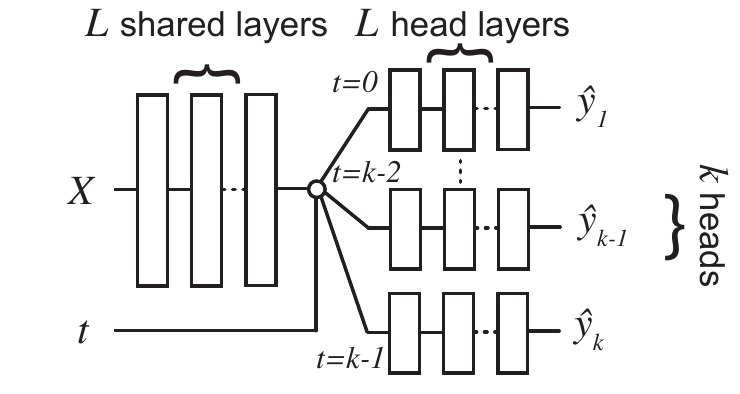}
\vskip -1.25ex
\caption{The TARNET architecture with $k$ heads for the multiple treatment setting. Each head predicts a potential outcome $\hat{y}_j$, and is only trained on samples that received the respective treatment.} 
\label{fig:tarnet}
\vskip -2ex
\end{wrapfigure}

\paragraph{Model Architecture.} The chosen architecture plays a key role in the performance of neural networks when attempting to learn representations for counterfactual inference \cite{shalit2016estimating,alaa2018limits}. \citet{shalit2016estimating} claimed that the na\"ive approach of appending the treatment index $t_j$ may perform poorly if $X$ is high-dimensional, because the influence of $t_j$ on the hidden layers may be lost during training. \citet{shalit2016estimating} subsequently introduced the TARNET architecture to rectify this issue. Since the original TARNET was limited to the binary treatment setting, we extended the TARNET architecture to the multiple treatment setting (Figure \ref{fig:tarnet}). We did so by using $k$ head networks, one for each treatment over a set of shared base layers, each with $L$ layers. In TARNET, the $j$th head network is only trained on samples from treatment $t_j$. The shared layers are trained on all samples. By using a head network for each treatment, we ensure $t_j$ maintains an appropriate degree of influence on the network output.

\paragraph{Perfect Match (PM).} We consider fully differentiable neural network models $\hat{f}$ optimised via minibatch stochastic gradient descent (SGD) to predict potential outcomes $\hat{Y}$ for a given sample $x$. To address the treatment assignment bias inherent in observational data, we propose to perform SGD in a space that approximates that of a randomised experiment using the concept of balancing scores. Under unconfoundedness assumptions, balancing scores have the property that the assignment to treatment is unconfounded given the balancing score \cite{rosenbaum1983propensity,hirano2004propensity,ho2007matching}. The conditional probability $p(t | X=x)$ of a given sample $x$ receiving a specific treatment $t$, also known as the propensity score \cite{rosenbaum1983propensity}, and the covariates $X$ themselves are prominent examples of balancing scores \cite{rosenbaum1983propensity,ho2007matching}. Using balancing scores, we can construct \textit{virtually randomised minibatches} that approximate the corresponding randomised experiment for the given counterfactual inference task by imputing, for each observed pair of covariates $x$ and factual outcome $y_t$, the remaining unobserved counterfactual outcomes by the outcomes of nearest neighbours in the training data by some balancing score, such as the propensity score. Formally, this approach is, when converged, equivalent to a nearest neighbour estimator for which we are guaranteed to have access to a perfect match, i.e. an exact match in the balancing score, for observed factual outcomes. We outline the Perfect Match (PM) algorithm in Algorithm \ref{alg:pm} (complexity analysis and implementation details in Appendix D). Upon convergence at the training data, neural networks trained using virtually randomised minibatches in the limit $N \to \infty$  remove any treatment assignment bias present in the data.
\begin{theorem}
\vskip -1.5ex
Upon convergence, under assumption (1) and for $N\to\infty$, a neural network $\hat{f}$ trained according to the PM algorithm is a consistent estimator of the true potential outcomes $Y$ for each $t$.
\vskip -2ex
\[ \lim_{N\to\infty} \E(\hat{f} (x) | X, T=t) = y_t \text{\hspace{5ex}(Proof in Appendix C)}\]
\end{theorem}
The optimal choice of balancing score for use in the PM algorithm depends on the properties of the dataset. For high-dimensional datasets, the scalar propensity score is preferable because it avoids the curse of dimensionality that would be associated with matching on the potentially high-dimensional $X$ directly. For low-dimensional datasets, the covariates $X$ are a good default choice as their use does not require a model of treatment propensity. The advantage of matching on the minibatch level, rather than the dataset level \cite{ho2011matchit}, is that it reduces the variance during training which in turn leads to better expected performance for counterfactual inference (Appendix E). In this sense, PM can be seen as a minibatch sampling strategy \cite{csiba2018importance} designed to improve learning for counterfactual inference. Propensity Dropout (PD) \cite{alaa2017deep} is another method using balancing scores that has been proposed to dynamically adjust the dropout regularisation strength for each observed sample depending on its treatment propensity. PD, in essence, discounts samples that are far from equal propensity for each treatment during training. This regularises the treatment assignment bias but also introduces data sparsity as not all available samples are leveraged equally for training. PM, in contrast, fully leverages all training samples by matching them with other samples with similar treatment propensities.

\begin{algorithm}
\caption{Perfect Match (PM). After augmentation, each batch contains an equal number of samples from each treatment and the covariates $x_i$ across treatments are approximately balanced.}
\label{alg:pm}
\begin{algorithmic}[1]
\Require Batch of $B$ random samples $X_\text{batch}$ with assigned treatments $t$, training set $X_\text{train}$ of $N$ samples, number of treatment options $k$, propensity score estimator $E_\text{PS}$ to calculate the probability $p(t|X)$ of treatment assigned given a sample $X$
\Ensure Batch $X_\text{out}$ consisting of $B\times k$ matched samples
\Procedure{Perfect\_Match:}{}
\State $X_\text{out} \gets$ Empty
\For {sample $X$ with treatment $t$ in $X_\text{batch}$}
\State $p(t|X) \gets E_\text{PS}(X)$
\For {$i=0$ to $k-1$}
\If{$i \neq t$}
\State $ps_i \gets p(t|X)_i$
\State $X_\text{matched} \gets$ get closest match to propensity score $ps_i$ with treatment $i$ from $X_\text{train}$ 
\State Add sample $X_\text{matched}$ to $X_\text{out}$
\EndIf
\EndFor
\State Add $X$ to $X_\text{out}$
\EndFor
\EndProcedure
\end{algorithmic}
\end{algorithm}
\setlength{\textfloatsep}{10pt}

\paragraph{Model Selection.} Besides accounting for the treatment assignment bias, the other major issue in learning for counterfactual inference from observational data is that, given multiple models, it is not trivial to decide which one to select. The root problem is that we do not have direct access to the true error in estimating counterfactual outcomes, only the error in estimating the observed factual outcomes. This makes it difficult to perform parameter and hyperparameter optimisation, as we are not able to evaluate which models are better than others for counterfactual inference on a given dataset. To rectify this problem, we use a nearest neighbour approximation $\hat\epsilon_\text{NN-PEHE}$ of the $\hat\epsilon_\text{PEHE}$ metric for the binary \cite{shalit2016estimating,schuler2018comparison} and multiple treatment settings for model selection. The $\hat\epsilon_\text{NN-PEHE}$ estimates the treatment effect of a given sample by substituting the true counterfactual outcome with the outcome $y_j$ from a respective nearest neighbour $\text{NN}$ matched on $X$ using the Euclidean distance.
\vskip -4.5ex
\begin{align}
\label{eq:NN_PEHE} \hat\epsilon_\text{NN-PEHE} = \frac{1}{N}\sum_{n=0}^{N}&\Big( [y_1(\text{NN}(n)) - y_0(\text{NN}(n))] - [\hat{y}_1(n) - \hat{y}_0(n)] \Big)^2
\end{align}
\vskip -1.85ex
Analogously to Equations (\ref{eq:multi_PEHE}) and (\ref{eq:multi_ATE}), the $\hat\epsilon_\text{NN-PEHE}$ metric can be extended to the multiple treatment setting by considering the mean $\hat\epsilon_\text{NN-PEHE}$ between all ${k \choose 2}$ possible pairs of treatments (Appendix F).

\section{Experiments}
Our experiments aimed to answer the following questions:
\begin{enumerate}[noitemsep,leftmargin=3.85ex]
\item[(1)] What is the comparative performance of PM in inferring counterfactual outcomes in the binary and multiple treatment setting compared to existing state-of-the-art methods?
\item[(2)] Does model selection by NN-PEHE outperform selection by factual MSE?
\item[(3)] How does the relative number of matched samples within a minibatch affect performance?
\item[(4)] How well does PM cope with an increasing treatment assignment bias in the observed data?
\item[(5)] How do the learning dynamics of minibatch matching compare to dataset-level matching?
\end{enumerate}

\subsection{Datasets}
We performed experiments on two real-world and semi-synthetic datasets with binary and multiple treatments in order to gain a better understanding of the empirical properties of PM. 

\paragraph{Infant Health and Development Program (IHDP).} The IHDP dataset \cite{hill2011bayesian} contains data from a randomised study on the impact of specialist visits on the cognitive development of children, and consists of 747 children with 25 covariates describing properties of the children and their mothers. Children that did not receive specialist visits were part of a control group. The outcomes were simulated using the NPCI package from \cite{dorie2016npci}\footnote{We used the same simulated outcomes as \citet{shalit2016estimating}.}. The IHDP dataset is biased because the treatment groups had a biased subset of the treated population removed \cite{shalit2016estimating}.

\begin{figure}[b!]
\vskip -3.25ex
\centering
\includegraphics[width=0.25\linewidth]{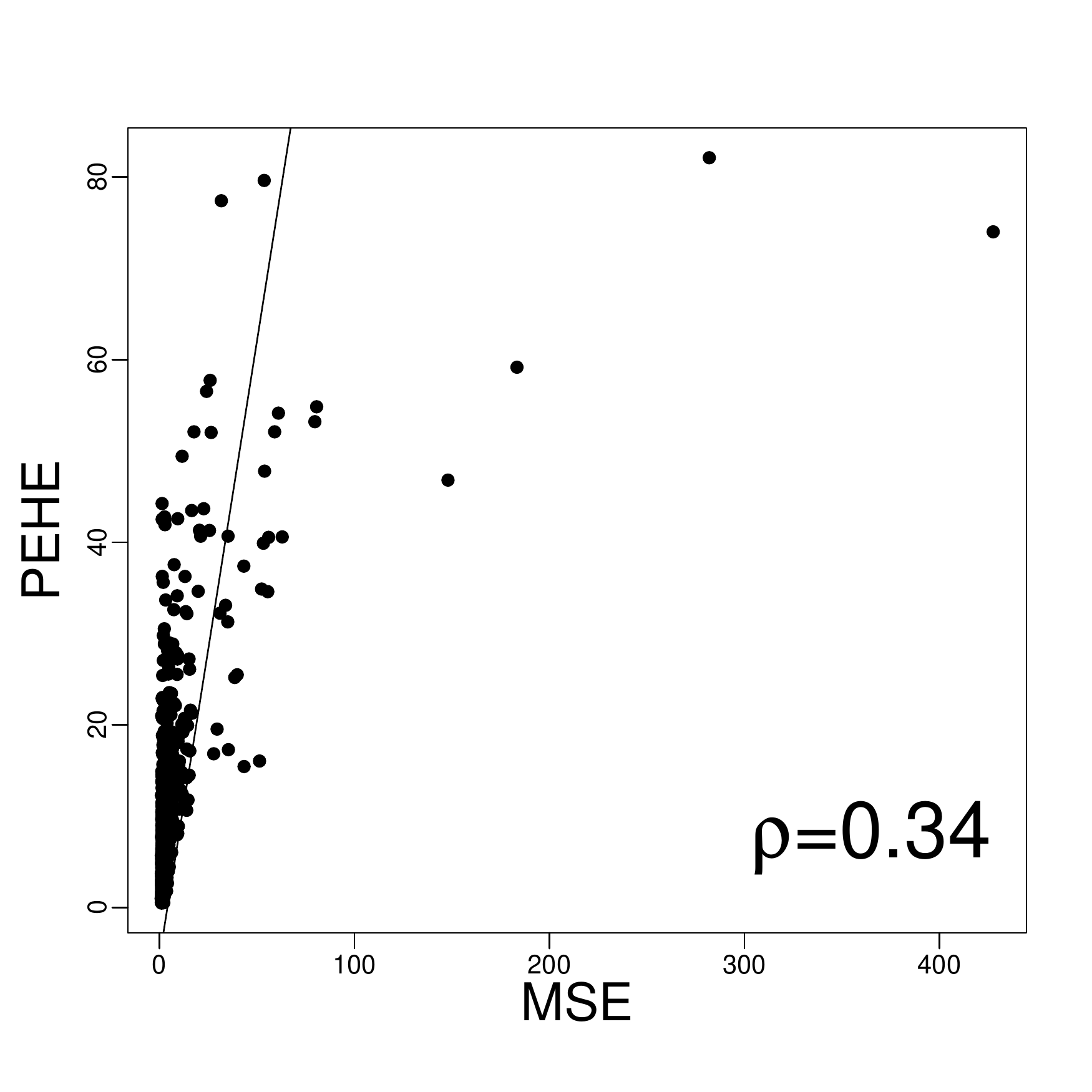}\hspace{12ex}
\includegraphics[width=0.25\linewidth]{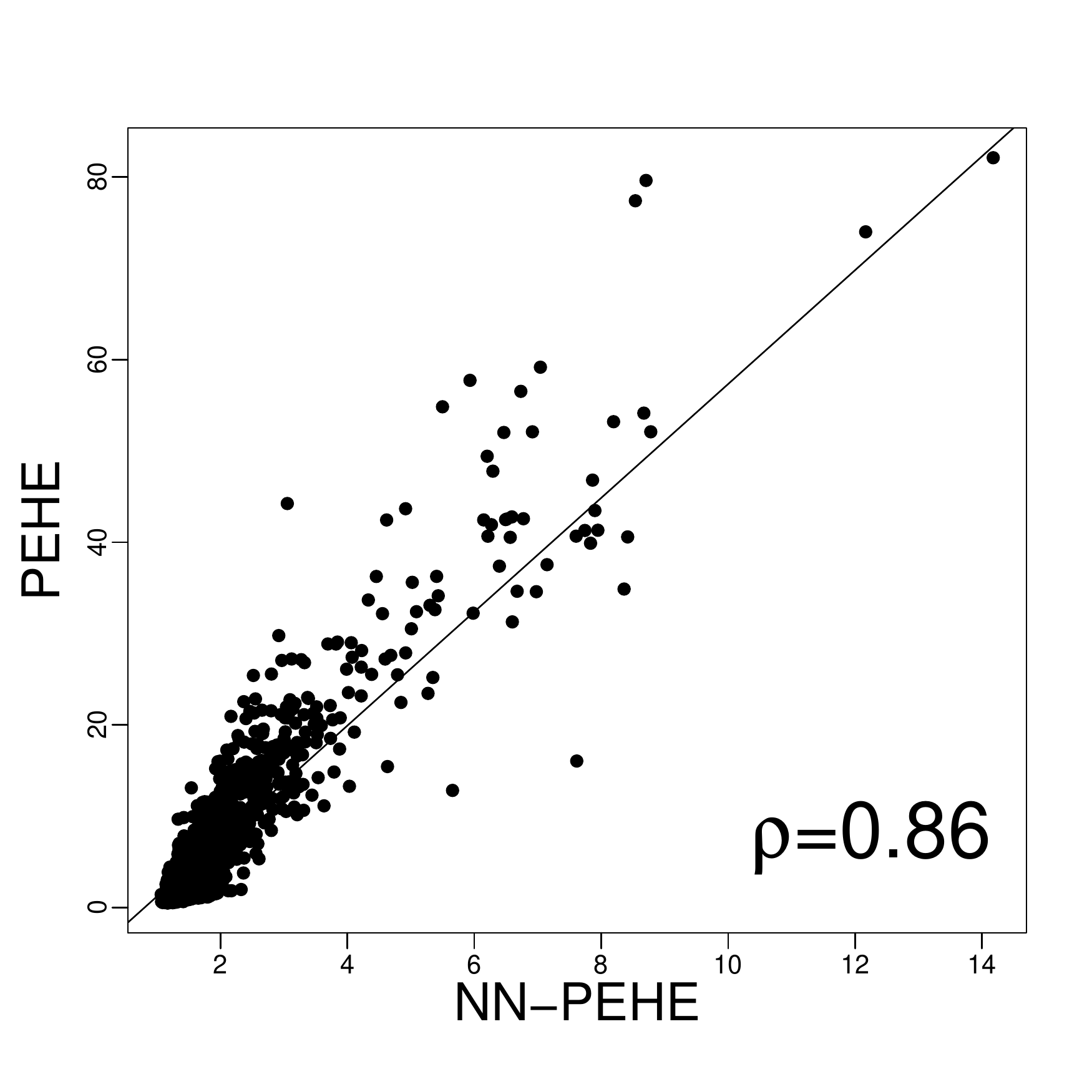}
\vskip -2ex
\caption{Correlation analysis of the real PEHE (y-axis) with the mean squared error (MSE; left) and the nearest neighbour approximation of the precision in estimation of heterogenous effect (NN-PEHE; right) across over 20000 model evaluations on the validation set of IHDP. Scatterplots show a subsample of 1400 data points. $\rho$ indicates the Pearson correlation.}
\label{fig:pehe_correlation}
\vskip -1.5ex
\end{figure}

\paragraph{News.} The News dataset was first proposed as a benchmark for counterfactual inference by \citet{johansson2016learning} and consists of 5000 randomly sampled news articles from the NY Times corpus\footnote{https://archive.ics.uci.edu/ml/datasets/bag+of+words}. The News dataset contains data on the opinion of media consumers on news items. The samples $X$ represent news items consisting of word counts $x_i \in \mathbb{N}$, the outcome $y_j \in \mathbb{R}$ is the reader's opinion of the news item, and the $k$ available treatments represent various devices that could be used for viewing, e.g. smartphone, tablet, desktop, television or others \cite{johansson2016learning}. We extended the original dataset specification in \cite{johansson2016learning} to enable the simulation of arbitrary numbers of viewing devices. To model that consumers prefer to read certain media items on specific viewing devices, we train a topic model on the whole NY Times corpus and define $z(X)$ as the topic distribution of news item $X$. We then randomly pick $k+1$ centroids in topic space, with $k$ centroids $z_j$ per viewing device and one control centroid $z_c$. We assigned a random Gaussian outcome distribution with mean $\mu_{j} \sim \mathcal{N}(0.45, 0.15)$ and standard deviation $\sigma_{j} \sim \mathcal{N}(0.1,0.05)$ to each centroid. For each sample, we drew ideal potential outcomes from that Gaussian outcome distribution $\tilde{y}_j \sim \mathcal{N}(\mu_j, \sigma_j) + \epsilon$ with $\epsilon \sim \mathcal{N}(0, 0.15)$. We then defined the unscaled potential outcomes $\bar{y_j} = \tilde{y}_j*[\text{D}(z(X), z_j) + \text{D}(z(X), z_c)]$ as the ideal potential outcomes $\tilde{y}_j$ weighted by the sum of distances to centroids $z_j$ and the control centroid $z_c$ using the Euclidean distance as distance $\text{D}$. We assigned the observed treatment $t$ using $t|x \sim \text{Bern}(\text{softmax}(\kappa \bar{y_j}))$ with a treatment assignment bias coefficient $\kappa$, and the true potential outcome $y_j = C\bar{y_j}$ as the unscaled potential outcomes $\bar{y_j}$ scaled by a  coefficient $C=50$. We used four different variants of this dataset with $k=2$, $4$, $8$, and $16$ viewing devices, and $\kappa = 10$, $10$, $10$, and $7$, respectively. Higher values of $\kappa$ indicate a higher expected assignment bias depending on $\bar{y_j}$. $\kappa = 0$ indicates no assignment bias. 

\begin{figure*}[t]
\centering
\includegraphics[height=23mm]{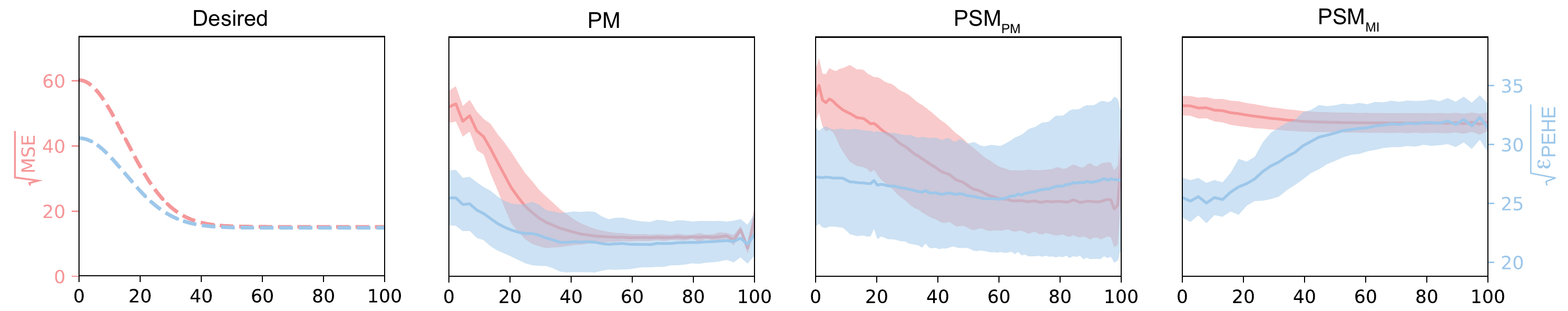}
\vskip -0.5ex
\caption{Comparison of the learning dynamics during training (normalised training epochs; from start = 0 to end = 100 of training, x-axis) of several matching-based methods on the validation set of News-8. The coloured lines correspond to the mean value of the factual error ($\sqrt{\text{MSE}}$; red) and the counterfactual error ($\sqrt{\hat\epsilon_\text{PEHE}}$; blue) across 10 randomised hyperparameter configurations (lower is better). The shaded area indicates the standard deviation. All methods used exactly the same model and hyperparameters and only differed in how they addressed treatment assignment bias. The leftmost figure shows the desired behavior of the counterfactual and factual error jointly decreasing until convergence. PM best matches the desired behavior of aligning factual and counterfactual error.} 
\label{fig:psm_comparison}
\vskip -3.5ex
\end{figure*}

All datasets with the exception of IHDP were split into a training (63\%), validation (27\%) and test set (10\% of samples). For IHDP we used exactly the same splits as previously used by \citet{shalit2016estimating}. We repeated experiments on IHDP and News 1000 and 50 times, respectively. We reassigned outcomes and treatments with a new random seed for each repetition. 

\setlength{\textfloatsep}{20.0pt plus 2.0pt minus 4.0pt}
\subsection{Experimental Setup}

\paragraph{Models.} We evaluated PM, ablations, baselines, and all relevant state-of-the-art methods: kNN \cite{ho2007matching}, BART \cite{chipman2010bart,chipman2016bayestree}, Random Forests (RF) \cite{breiman2001random}, CF \cite{wager2017estimation}, GANITE \cite{yoon2018ganite}, Balancing Neural Network (BNN) \cite{johansson2016learning},  TARNET \cite{shalit2016estimating},  Counterfactual Regression Network using the Wasserstein regulariser (CFRNET$_\text{Wass}$) \cite{shalit2016estimating}, and PD \cite{alaa2017deep}. We trained a Support Vector Machine (SVM) with probability estimation \cite{scikitlearn} to estimate $p(t | X)$ for PM on the training set. We also evaluated preprocessing the entire training set with PSM using the same matching routine as PM (PSM$_\text{PM}$) and the "MatchIt" package (PSM$_\text{MI}$, \cite{ho2011matchit} before training a TARNET (Appendix G). In addition, we trained an ablation of PM where we matched on the covariates $X$ (+ on $X$) directly, if $X$ was low-dimensional ($p<200$), and on a 50-dimensional representation of $X$ obtained via principal components analysis (PCA), if $X$ was high-dimensional, instead of on the propensity score. We also evaluated PM with a multi-layer perceptron (+ MLP) that received the treatment index $t_j$ as an input instead of using a TARNET.

\begin{figure}[b!]
\vskip -1ex
\centering
\begin{minipage}[b]{.49\textwidth}
 \centering
  \includegraphics[width=0.98\linewidth]{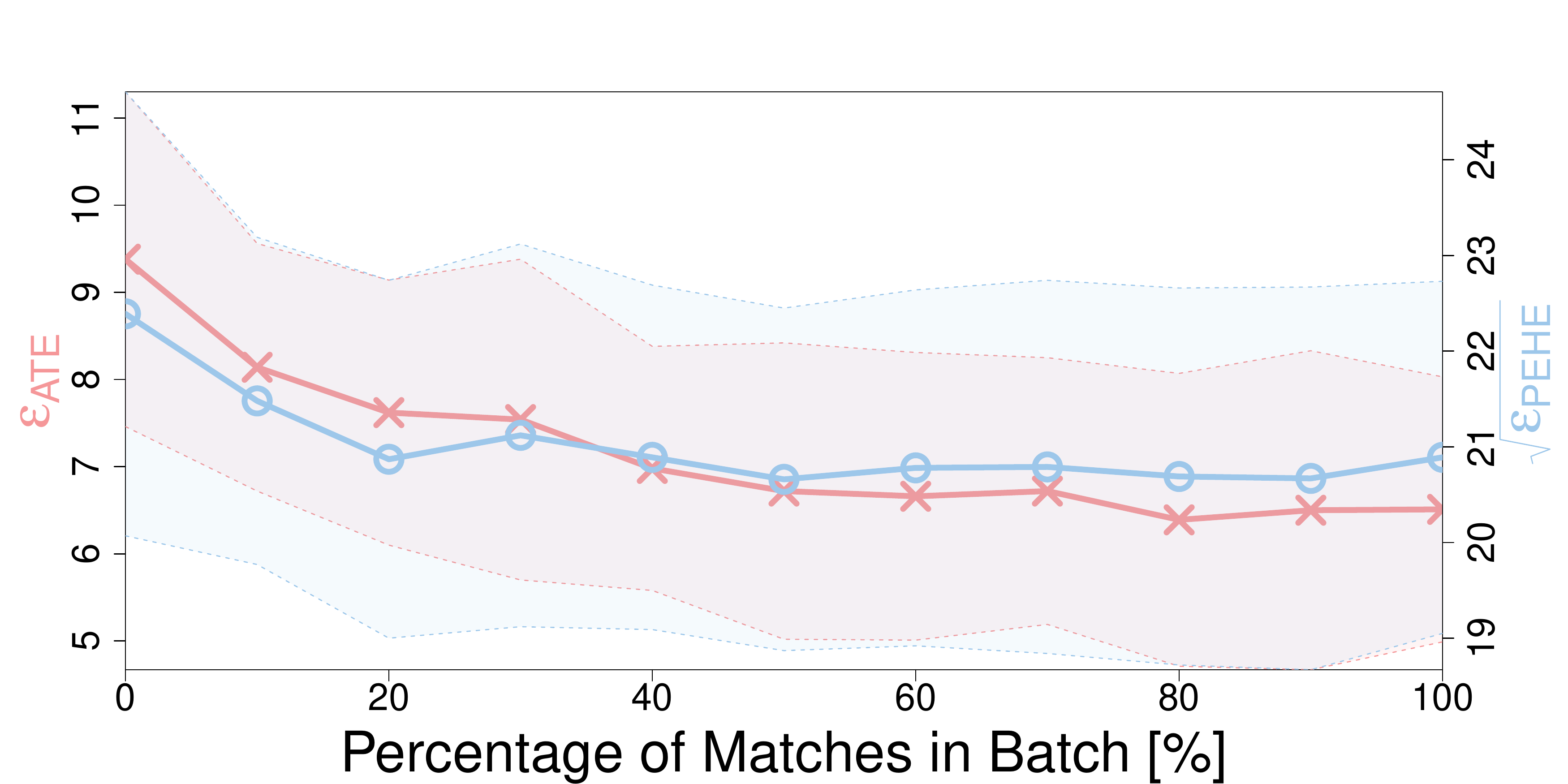}
  \vskip -1ex
  \captionof{figure}{Change in error (y-axes) in terms of precision in estimation of heterogenous effect (PEHE) and average treatment effect (ATE) when increasing the percentage of matches in each minibatch (x-axis). Symbols correspond to the mean value of $\hat\epsilon_\text{mATE}$ (red) and $\sqrt{\hat\epsilon_\text{mPEHE}}$ (blue) on the test set of News-8 across 50 repeated runs with new outcomes (lower is better). } 
  \label{fig:matches}
\end{minipage}
\hfill
\begin{minipage}[b]{.49\textwidth}
  \centering
  \includegraphics[width=0.98\linewidth]{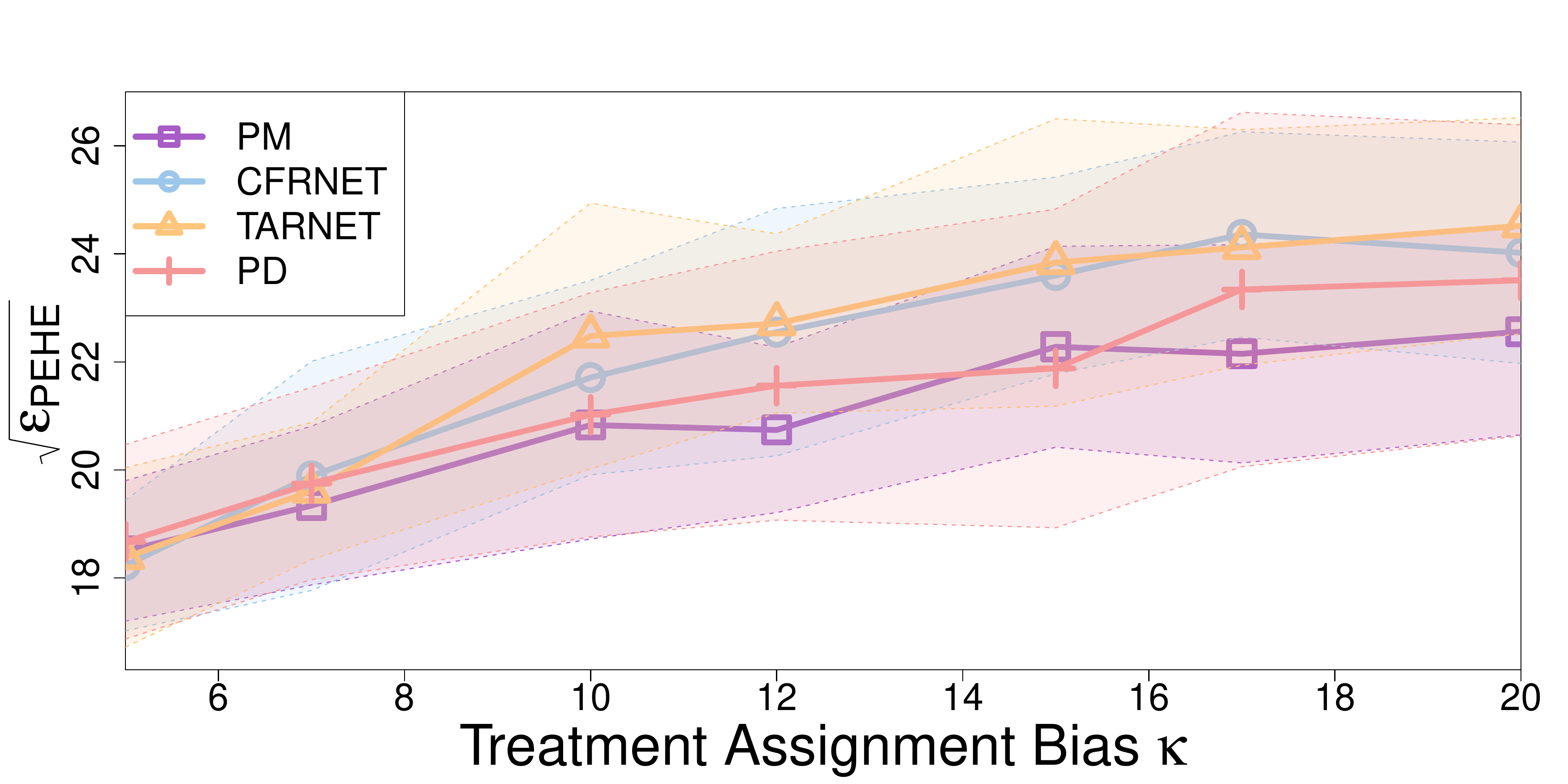}
  \vskip -1ex
  \captionof{figure}{Comparison of several state-of-the-art methods for counterfactual inference on the test set of the News-8 dataset when varying the treatment assignment imbalance $\kappa$ (x-axis), i.e. how much the treatment assignment is biased towards more effective treatments. Symbols correspond to the mean value of $\sqrt{\hat\epsilon_\text{mPEHE}}$ across 50 repeated runs with new outcomes (lower is better). }    \label{fig:assignment_bias}
\end{minipage}
\end{figure}

\paragraph{Architectures.} To ensure that differences between methods of learning counterfactual representations for neural networks are not due to differences in architecture, we based the neural architectures for TARNET, CFRNET$_\text{Wass}$, PD and PM on the same, previously described extension of the TARNET architecture \cite{shalit2016estimating} (Appendix H) to the multiple treatment setting.

\paragraph{Hyperparameters.} For the IHDP and News datasets we respectively used 30 and 10 optimisation runs for each method using randomly selected hyperparameters from predefined ranges (Appendix I). We selected the best model across the runs based on validation set $\hat\epsilon_\text{NN-PEHE}$ or $\hat\epsilon_\text{NN-mPEHE}$. 

\paragraph{Metrics.} We calculated the $\epsilon_\text{PEHE}$ (Eq. \ref{eq:PEHE}) and $\epsilon_\text{ATE}$ (Appendix B) for the binary IHDP and News-2 datasets, and the $\hat\epsilon_\text{mPEHE}$ (Eq. \ref{eq:multi_PEHE}) and $\hat\epsilon_\text{mATE}$ (Eq. \ref{eq:multi_ATE}) for News-4/8/16 datasets. 

\section{Results and Discussion}
\paragraph{Counterfactual Inference.} We evaluated the counterfactual inference performance of the listed models in settings with two or more available treatments (Table \ref{tb:results_binary}, ATEs in Appendix Table S3). On IHDP, the PM variants reached the best performance in terms of $\sqrt{\epsilon_\text{PEHE}}$, and the second best ${\epsilon_\text{ATE}}$ after CFRNET. On the binary News-2, PM outperformed all other methods in terms of $\sqrt{\epsilon_\text{PEHE}}$ and ${\epsilon_\text{ATE}}$. On the News-4/8/16 datasets with more than two treatments, PM consistently outperformed all other methods - in some cases by a large margin - on both metrics with the exception of the News-4 dataset, where PM came second to PD. The strong performance of PM across a wide range of datasets with varying amounts of treatments is remarkable considering how simple it is compared to other, highly specialised methods. Notably, PM consistently outperformed both CFRNET, which accounted for covariate imbalances between treatments via regularisation rather than matching, and PSM$_\text{MI}$, which accounted for covariate imbalances by preprocessing the entire training set with a matching algorithm \cite{ho2011matchit}. We also found that matching on the propensity score was, in almost all cases, not significantly different from matching on $X$ directly when $X$ was low-dimensional, or a low-dimensional representation of $X$ when $X$ was high-dimensional (+ on $X$). This indicates that PM is effective with any low-dimensional balancing score. In addition, using PM with the TARNET architecture outperformed the MLP (+ MLP) in almost all cases, with the exception of the low-dimensional IHDP. We therefore conclude that matching on the propensity score or a low-dimensional representation of $X$ and using the TARNET architecture are sensible default configurations, particularly when $X$ is high-dimensional. Finally, although TARNETs trained with PM have similar asymptotic properties as kNN, we found that TARNETs trained with PM significantly outperformed kNN in all cases. This is likely due to the shared base layers that enable them to efficiently share information across the per-treatment representations in the head networks.

\paragraph{Model Selection.} To judge whether NN-PEHE is more suitable for model selection for counterfactual inference than MSE, we compared their respective correlations with the PEHE on IHDP. We found that NN-PEHE correlates significantly better with the PEHE than MSE (Figure \ref{fig:pehe_correlation}).

\paragraph{Number of Matches per Minibatch.} To determine the impact of matching fewer than 100\% of all samples in a batch, we evaluated PM on News-8 trained with varying percentages of matched samples on the range 0 to 100\% in steps of 10\% (Figure \ref{fig:matches}). We found that including more matches indeed consistently reduces the counterfactual error up to 100\% of samples matched. Interestingly, we found a large improvement over using no matched samples even for relatively small percentages ($<$40\%) of matched samples per batch. This shows that propensity score matching within a batch is indeed  effective at improving the training of neural networks for counterfactual inference.

\paragraph{Treatment Assignment Bias.} To assess how the predictive performance of the different methods is influenced by increasing amounts of treatment assignment bias, we evaluated their performances on News-8 while varying the assignment bias coefficient $\kappa$ on the range of 5 to 20 (Figure \ref{fig:assignment_bias}). We found that PM handles high amounts of assignment bias better than existing state-of-the-art methods.

\begin{table*}[t!]
\setlength{\tabcolsep}{1.55ex}
\caption{Comparison of methods for counterfactual inference with two and more available treatments on IHDP and News-2/4/8/16. We report the mean value $\pm$ the standard deviation of $\sqrt{\epsilon_\text{PEHE}}$, $\sqrt{\hat\epsilon_\text{PEHE}}$, and $\sqrt{\hat\epsilon_\text{mPEHE}}$ on the test sets over 1000 and 50 repeated runs for IHDP and News-2/4/8/16, respectively. Best results on each benchmark in bold. \textsuperscript{$\dagger$} = significantly different from PM ($t$-test, $\alpha < 0.05$).}
\label{tb:results_binary}
\centering
\begin{small}
\begin{tabular}{l@{\hskip 3.5ex}r@{\hskip 3.5ex}r@{\hskip 3.5ex}r@{\hskip 3.5ex}r@{\hskip 3.5ex}r}
\toprule
& IHDP & News-2 & News-4 & News-8 & News-16 \\
Method & $\sqrt{\epsilon_\text{PEHE}}$ & $\sqrt{\hat\epsilon_\text{PEHE}}$ & $\sqrt{\hat\epsilon_\text{mPEHE}}$ & $\sqrt{\hat\epsilon_\text{mPEHE}}$ & $\sqrt{\hat\epsilon_\text{mPEHE}}$ \\
\midrule
{PM }\hspace{5ex}   &  {\hspace{1ex}0.84} $\pm$ 0.61 & \textbf{16.76} $\pm$ 1.26 & {21.58} $\pm$ 2.58 & \textbf{20.76} $\pm$ 1.86 & \textbf{20.24} $\pm$ 1.46  \\
+ on $X$  &   \textbf{0.81} $\pm$ 0.57 & {17.06} $\pm$ 1.22 & {21.41} $\pm$ 1.75 & {20.90} $\pm$ 2.07 & {20.67} $\pm$ 1.42  \\
+ MLP  &  {0.83} $\pm$ 0.57 & \textsuperscript{$\dagger$} {18.38} $\pm$ 1.46 & \textsuperscript{$\dagger$} {25.05} $\pm$ 2.80 & \textsuperscript{$\dagger$} {24.88} $\pm$ 1.98 & \textsuperscript{$\dagger$}  {27.05} $\pm$ 2.47  \\
\midrule
kNN  &  \textsuperscript{$\dagger$} {6.66} $\pm$ 6.89 & \textsuperscript{$\dagger$} {18.14} $\pm$ 1.64 & \textsuperscript{$\dagger$} {27.92} $\pm$ 2.44 & \textsuperscript{$\dagger$} {26.20} $\pm$ 2.18 &  \textsuperscript{$\dagger$} {27.64} $\pm$ 2.40  \\
PSM$_\text{PM}$  &  \textsuperscript{$\dagger$} {2.36} $\pm$ 3.39 & \textsuperscript{$\dagger$} {17.49} $\pm$ 1.49 & \textsuperscript{$\dagger$} {22.74} $\pm$ 2.58 & \textsuperscript{$\dagger$} {22.16} $\pm$ 1.79 & \textsuperscript{$\dagger$} {23.57} $\pm$ 2.48  \\
PSM$_\text{MI}$  &  \textsuperscript{$\dagger$} {2.70} $\pm$ 3.85 & \textsuperscript{$\dagger$} {17.40} $\pm$ 1.30 & \textsuperscript{$\dagger$} {37.26} $\pm$ 2.28 & \textsuperscript{$\dagger$} {30.50} $\pm$ 1.70 &  \textsuperscript{$\dagger$} {28.17} $\pm$ 2.02  \\
\midrule
RF  &  \textsuperscript{$\dagger$} {4.54} $\pm$ 7.09 & \textsuperscript{$\dagger$} {17.39} $\pm$ 1.24 & \textsuperscript{$\dagger$} {26.59} $\pm$ 3.02 & \textsuperscript{$\dagger$} {23.77} $\pm$ 2.14 & \textsuperscript{$\dagger$}  {26.13} $\pm$ 2.48  \\
CF  &  \textsuperscript{$\dagger$} {4.47} $\pm$ 6.55 & \textsuperscript{$\dagger$} {17.59} $\pm$ 1.63 & \textsuperscript{$\dagger$} {23.86} $\pm$ 2.50 & \textsuperscript{$\dagger$} {22.56} $\pm$ 2.32 & \textsuperscript{$\dagger$}  {21.45} $\pm$ 2.23 \\
\midrule
BART  &  \textsuperscript{$\dagger$} {2.57} $\pm$ 3.97 & \textsuperscript{$\dagger$} {18.53} $\pm$ 2.02 & \textsuperscript{$\dagger$} {26.41} $\pm$ 3.10 & \textsuperscript{$\dagger$} {25.78} $\pm$ 2.66 &  \textsuperscript{$\dagger$} {27.45} $\pm$ 2.84  \\
GANITE  &  \textsuperscript{$\dagger$} {5.79} $\pm$ 8.35 & \textsuperscript{$\dagger$} {18.28} $\pm$ 1.66 & \textsuperscript{$\dagger$} {24.50} $\pm$ 2.27 & \textsuperscript{$\dagger$} {23.58} $\pm$ 2.48 &  \textsuperscript{$\dagger$} {25.12} $\pm$ 3.53  \\
PD  &  \textsuperscript{$\dagger$} {5.14} $\pm$ 6.55 & \textsuperscript{$\dagger$} {17.52} $\pm$ 1.62 & \textbf{20.88} $\pm$ 3.24 & {21.19} $\pm$ 2.29 &  \textsuperscript{$\dagger$} {22.28} $\pm$ 2.25  \\
TARNET  &  \textsuperscript{$\dagger$} {1.32} $\pm$ 1.61 & {17.17} $\pm$ 1.25 & \textsuperscript{$\dagger$} {23.40} $\pm$ 2.20 & \textsuperscript{$\dagger$} {22.39} $\pm$ 2.32 &  \textsuperscript{$\dagger$} {21.19} $\pm$ 2.01  \\
CFRNET$_{\text{Wass}}$  &  {0.88} $\pm$ 1.25 & {16.93} $\pm$ 1.12 & \textsuperscript{$\dagger$} {22.65} $\pm$ 1.97 & \textsuperscript{$\dagger$} {21.64} $\pm$ 1.82 & \textsuperscript{$\dagger$} {20.87} $\pm$ 1.46  \\
\bottomrule
\end{tabular}
\end{small}
\vskip -4.5ex
\end{table*}

\paragraph{Comparing Minibatch and Dataset Matching.} As outlined previously, if we were successful in balancing the covariates using the balancing score, we would expect that the counterfactual error is implicitly and consistently improved alongside the factual error. To elucidate to what degree this is the case when using the matching-based methods we compared, we evaluated the respective training dynamics of PM, PSM$_\text{PM}$ and PSM$_\text{MI}$ (Figure \ref{fig:psm_comparison}). We found that PM better conforms to the desired behavior than PSM$_\text{PM}$ and PSM$_\text{MI}$. PSM$_\text{PM}$, which used the same matching strategy as PM but on the dataset level, showed a much higher variance than PM. PSM$_\text{MI}$ was overfitting to the treated group.

\paragraph{Limitations.} A general limitation of this work, and most related approaches, to counterfactual inference from observational data is that its underlying theory only holds under the assumption that there are no unobserved confounders - which guarantees identifiability of the causal effects. However, it has been shown that hidden confounders may not necessarily decrease the performance of ITE estimators in practice if we observe suitable proxy variables \cite{montgomery2000measuring,louizos2017causal}. 

\section{Conclusion}
We presented PM, a new and simple method for training neural networks for estimating ITEs from observational data that extends to any number of available treatments. In addition, we extended the TARNET architecture and the PEHE metric to settings with more than two treatments, and introduced a nearest neighbour approximation of PEHE and mPEHE that can be used for model selection without having access to counterfactual outcomes. We performed experiments on several real-world and semi-synthetic datasets that showed that PM outperforms a number of more complex state-of-the-art methods in inferring counterfactual outcomes. We also found that the NN-PEHE correlates significantly better with real PEHE than MSE, that including more matched samples in each minibatch improves the learning of counterfactual representations, and that PM handles an increasing treatment assignment bias better than existing state-of-the-art methods. PM may be used for settings with any amount of treatments, is compatible with any existing neural network architecture, simple to implement, and does not introduce any additional hyperparameters or computational complexity. Flexible and expressive models for learning counterfactual representations that generalise to settings with multiple available treatments could potentially facilitate the derivation of valuable insights from observational data in several important domains, such as healthcare, economics and public policy.

\nocite{kapelner2013bartmachine,athey2016generalized}

\subsubsection*{Acknowledgments}
This work was partially funded by the Swiss National Science Foundation (SNSF) project No. 167302 within the National Research Program (NRP) $75$ ``Big Data''. We gratefully acknowledge the support of NVIDIA Corporation with the donation of the Titan Xp GPUs used for this research. 

\bibliography{references}

\begin{thebibliography}{35}
\providecommand{\natexlab}[1]{#1}
\providecommand{\url}[1]{\texttt{#1}}
\expandafter\ifx\csname urlstyle\endcsname\relax
  \providecommand{\doi}[1]{doi: #1}\else
  \providecommand{\doi}{doi: \begingroup \urlstyle{rm}\Url}\fi

\bibitem[Shalit et~al.(2017)Shalit, Johansson, and
  Sontag]{shalit2016estimating}
Uri Shalit, Fredrik~D Johansson, and David Sontag.
\newblock {Estimating individual treatment effect: Generalization bounds and
  algorithms}.
\newblock In \emph{{International Conference on Machine Learning}}, 2017.

\bibitem[LaLonde(1986)]{lalonde1986evaluating}
Robert~J LaLonde.
\newblock Evaluating the econometric evaluations of training programs with
  experimental data.
\newblock \emph{The American economic review}, pages 604--620, 1986.

\bibitem[Carpenter(2014)]{carpenter2014reputation}
Daniel Carpenter.
\newblock \emph{Reputation and power: organizational image and pharmaceutical
  regulation at the FDA}.
\newblock Princeton University Press, 2014.

\bibitem[Bothwell et~al.(2016)Bothwell, Greene, Podolsky, and
  Jones]{bothwell2016rct}
Laura~E. Bothwell, Jeremy~A. Greene, Scott~H. Podolsky, and David~S. Jones.
\newblock {Assessing the Gold Standard — Lessons from the History of RCTs}.
\newblock \emph{New England Journal of Medicine}, 374\penalty0 (22):\penalty0
  2175--2181, 2016.

\bibitem[Ho et~al.(2007)Ho, Imai, King, and Stuart]{ho2007matching}
Daniel~E Ho, Kosuke Imai, Gary King, and Elizabeth~A Stuart.
\newblock Matching as nonparametric preprocessing for reducing model dependence
  in parametric causal inference.
\newblock \emph{Political analysis}, 15\penalty0 (3):\penalty0 199--236, 2007.

\bibitem[Indyk and Motwani(1998)]{indyk1998approximate}
Piotr Indyk and Rajeev Motwani.
\newblock Approximate nearest neighbors: towards removing the curse of
  dimensionality.
\newblock In \emph{Proceedings of the thirtieth annual ACM symposium on Theory
  of computing}, pages 604--613. ACM, 1998.

\bibitem[Rosenbaum and Rubin(1983)]{rosenbaum1983propensity}
Paul~R. Rosenbaum and Donald~B. Rubin.
\newblock The central role of the propensity score in observational studies for
  causal effects.
\newblock \emph{Biometrika}, 70\penalty0 (1):\penalty0 41--55, 1983.

\bibitem[Kallus(2017)]{kallus2017recursive}
Nathan Kallus.
\newblock Recursive partitioning for personalization using observational data.
\newblock In \emph{International Conference on Machine Learning}, 2017.

\bibitem[Funk et~al.(2011)Funk, Westreich, Wiesen, Stürmer, Brookhart, and
  Davidian]{funk2011doubly}
Michele~Jonsson Funk, Daniel Westreich, Chris Wiesen, Til Stürmer, M.~Alan
  Brookhart, and Marie Davidian.
\newblock Doubly robust estimation of causal effects.
\newblock \emph{American Journal of Epidemiology}, 173\penalty0 (7):\penalty0
  761--767, 2011.

\bibitem[Chipman et~al.(2010)Chipman, George, McCulloch,
  et~al.]{chipman2010bart}
Hugh~A Chipman, Edward~I George, Robert~E McCulloch, et~al.
\newblock {BART: Bayesian additive regression trees}.
\newblock \emph{The Annals of Applied Statistics}, 4\penalty0 (1):\penalty0
  266--298, 2010.

\bibitem[Chipman and McCulloch(2016)]{chipman2016bayestree}
Hugh Chipman and Robert McCulloch.
\newblock {BayesTree: Bayesian additive regression trees}.
\newblock \emph{{R package version 0.3-1.4}}, 2016.

\bibitem[Wager and Athey(2017)]{wager2017estimation}
Stefan Wager and Susan Athey.
\newblock Estimation and inference of heterogeneous treatment effects using
  random forests.
\newblock \emph{{Journal of the American Statistical Association}}, 2017.

\bibitem[Johansson et~al.(2016)Johansson, Shalit, and
  Sontag]{johansson2016learning}
Fredrik Johansson, Uri Shalit, and David Sontag.
\newblock Learning representations for counterfactual inference.
\newblock In \emph{International Conference on Machine Learning}, pages
  3020--3029, 2016.

\bibitem[Mansour et~al.(2009)Mansour, Mohri, and
  Rostamizadeh]{mansour2009domain}
Yishay Mansour, Mehryar Mohri, and Afshin Rostamizadeh.
\newblock Domain adaptation: Learning bounds and algorithms.
\newblock \emph{arXiv preprint arXiv:0902.3430}, 2009.

\bibitem[Alaa et~al.(2017)Alaa, Weisz, and van~der Schaar]{alaa2017deep}
Ahmed~M Alaa, Michael Weisz, and Mihaela van~der Schaar.
\newblock Deep counterfactual networks with propensity-dropout.
\newblock \emph{arXiv preprint arXiv:1706.05966}, 2017.

\bibitem[Yoon et~al.(2018)Yoon, Jordon, and van~der Schaar]{yoon2018ganite}
Jinsung Yoon, James Jordon, and Mihaela van~der Schaar.
\newblock {GANITE: Estimation of Individualized Treatment Effects using
  Generative Adversarial Nets}.
\newblock In \emph{{International Conference on Learning Representations}},
  2018.

\bibitem[Alaa and van~der Schaar(2017)]{alaa2017bayesian}
Ahmed~M Alaa and Mihaela van~der Schaar.
\newblock Bayesian inference of individualized treatment effects using
  multi-task gaussian processes.
\newblock In \emph{{Advances in Neural Information Processing Systems}}, pages
  3424--3432, 2017.

\bibitem[Rubin(2005)]{rubin2005causal}
Donald~B Rubin.
\newblock Causal inference using potential outcomes: Design, modeling,
  decisions.
\newblock \emph{Journal of the American Statistical Association}, 100\penalty0
  (469):\penalty0 322--331, 2005.

\bibitem[Pearl(2009)]{pearl2009causality}
Judea Pearl.
\newblock \emph{Causality}.
\newblock Cambridge university press, 2009.

\bibitem[Peters et~al.(2017)Peters, Janzing, and
  Sch{\"o}lkopf]{peters2017elements}
Jonas Peters, Dominik Janzing, and Bernhard Sch{\"o}lkopf.
\newblock \emph{Elements of causal inference: foundations and learning
  algorithms}.
\newblock MIT press, 2017.

\bibitem[Imbens(2000)]{imbens2000role}
Guido~W Imbens.
\newblock The role of the propensity score in estimating dose-response
  functions.
\newblock \emph{Biometrika}, 87\penalty0 (3):\penalty0 706--710, 2000.

\bibitem[Lechner(2001)]{lechner2001identification}
Michael Lechner.
\newblock Identification and estimation of causal effects of multiple
  treatments under the conditional independence assumption.
\newblock In \emph{{Econometric Evaluation of Labour Market Policies}}, pages
  43--58. Springer, 2001.

\bibitem[Hill(2011)]{hill2011bayesian}
Jennifer~L Hill.
\newblock Bayesian nonparametric modeling for causal inference.
\newblock \emph{Journal of Computational and Graphical Statistics}, 20\penalty0
  (1):\penalty0 217--240, 2011.

\bibitem[Alaa and Schaar(2018)]{alaa2018limits}
Ahmed Alaa and Mihaela Schaar.
\newblock Limits of estimating heterogeneous treatment effects: Guidelines for
  practical algorithm design.
\newblock In \emph{International Conference on Machine Learning}, pages
  129--138, 2018.

\bibitem[Hirano and Imbens(2004)]{hirano2004propensity}
Keisuke Hirano and Guido~W Imbens.
\newblock The propensity score with continuous treatments.
\newblock \emph{{Applied Bayesian Modeling and Causal Inference from
  Incomplete-data Perspectives}}, 226164:\penalty0 73--84, 2004.

\bibitem[Ho et~al.(2011)Ho, Imai, King, Stuart, et~al.]{ho2011matchit}
Daniel~E Ho, Kosuke Imai, Gary King, Elizabeth~A Stuart, et~al.
\newblock {MatchIt: nonparametric preprocessing for parametric causal
  inference}.
\newblock \emph{{Journal of Statistical Software}}, 42\penalty0 (8):\penalty0
  1--28, 2011.

\bibitem[Csiba and Richt{\'a}rik(2018)]{csiba2018importance}
Dominik Csiba and Peter Richt{\'a}rik.
\newblock Importance sampling for minibatches.
\newblock \emph{Journal of Machine Learning Research}, 19\penalty0 (27), 2018.

\bibitem[Schuler et~al.(2018)Schuler, Baiocchi, Tibshirani, and
  Shah]{schuler2018comparison}
Alejandro Schuler, Michael Baiocchi, Robert Tibshirani, and Nigam Shah.
\newblock A comparison of methods for model selection when estimating
  individual treatment effects.
\newblock \emph{arXiv preprint arXiv:1804.05146}, 2018.

\bibitem[Dorie(2016)]{dorie2016npci}
Vincent Dorie.
\newblock {NPCI: Non-parametrics for causal inference}, 2016.
\newblock URL \url{{https://github.com/vdorie/npci}}.

\bibitem[Breiman(2001)]{breiman2001random}
Leo Breiman.
\newblock Random forests.
\newblock \emph{{Machine learning}}, 45\penalty0 (1):\penalty0 5--32, 2001.

\bibitem[Pedregosa et~al.(2011)Pedregosa, Varoquaux, Gramfort, Michel, Thirion,
  Grisel, Blondel, Prettenhofer, Weiss, Dubourg, Vanderplas, Passos,
  Cournapeau, Brucher, Perrot, and Duchesnay]{scikitlearn}
F.~Pedregosa, G.~Varoquaux, A.~Gramfort, V.~Michel, B.~Thirion, O.~Grisel,
  M.~Blondel, P.~Prettenhofer, R.~Weiss, V.~Dubourg, J.~Vanderplas, A.~Passos,
  D.~Cournapeau, M.~Brucher, M.~Perrot, and E.~Duchesnay.
\newblock {Scikit-learn: Machine Learning in {P}ython}.
\newblock \emph{{Journal of Machine Learning Research}}, 12:\penalty0
  2825--2830, 2011.

\bibitem[Montgomery et~al.(2000)Montgomery, Gragnolati, Burke, and
  Paredes]{montgomery2000measuring}
Mark~R Montgomery, Michele Gragnolati, Kathleen~A Burke, and Edmundo Paredes.
\newblock Measuring living standards with proxy variables.
\newblock \emph{Demography}, 37\penalty0 (2):\penalty0 155--174, 2000.

\bibitem[Louizos et~al.(2017)Louizos, Shalit, Mooij, Sontag, Zemel, and
  Welling]{louizos2017causal}
Christos Louizos, Uri Shalit, Joris~M Mooij, David Sontag, Richard Zemel, and
  Max Welling.
\newblock Causal effect inference with deep latent-variable models.
\newblock In \emph{{Advances in Neural Information Processing Systems}}, pages
  6446--6456, 2017.

\bibitem[Kapelner and Bleich(2013)]{kapelner2013bartmachine}
Adam Kapelner and Justin Bleich.
\newblock {bartMachine: Machine learning with Bayesian additive regression
  trees}.
\newblock \emph{arXiv preprint arXiv:1312.2171}, 2013.

\bibitem[Athey et~al.(2016)Athey, Tibshirani, and Wager]{athey2016generalized}
Susan Athey, Julie Tibshirani, and Stefan Wager.
\newblock Generalized random forests.
\newblock \emph{arXiv preprint arXiv:1610.01271}, 2016.

\end{thebibliography}
\bibliographystyle{unsrtnat}

\includepdf[pages={1-}]{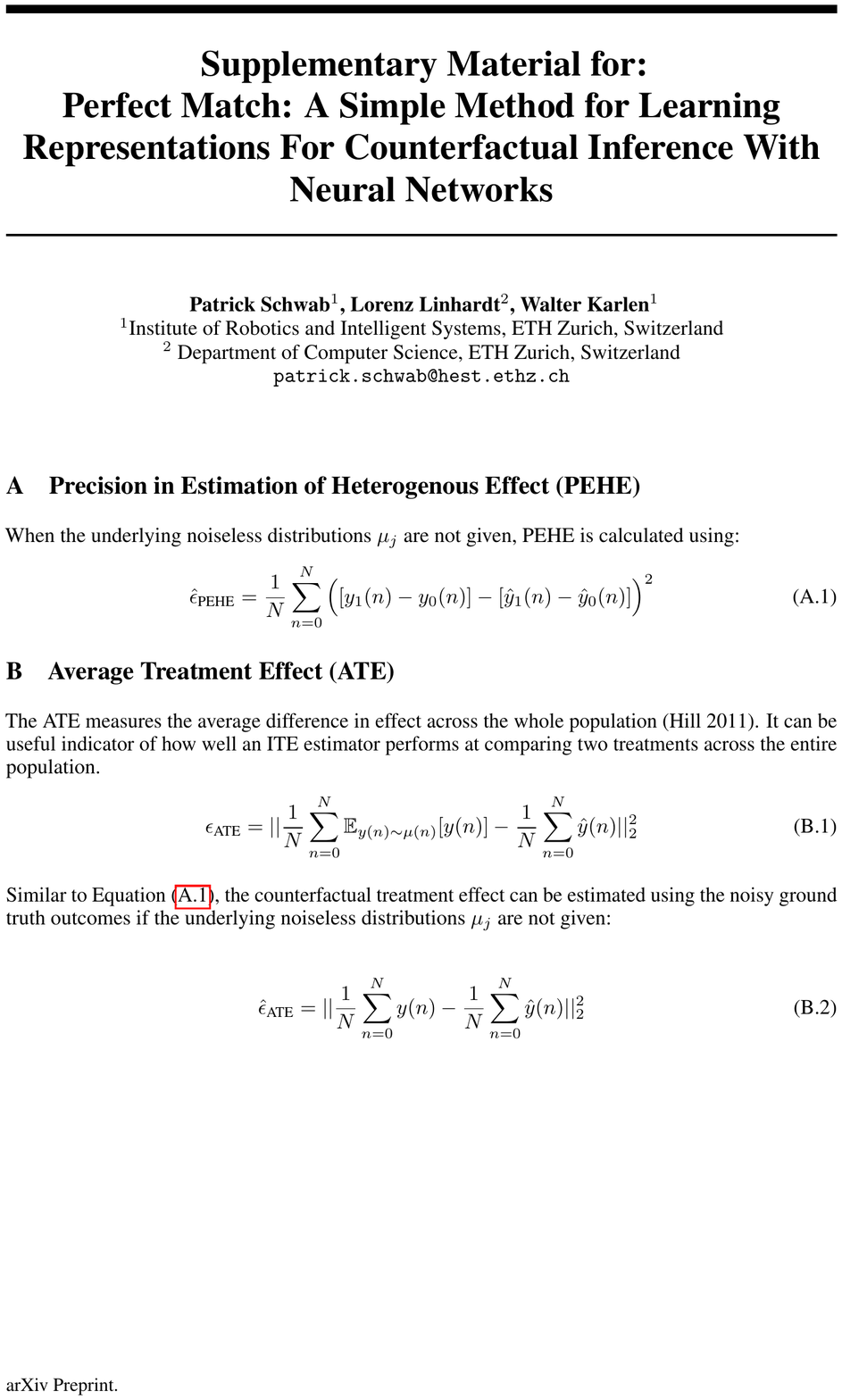}

\end{document}